\def\year{2021}\relax
\documentclass[11pt]{article} 

\usepackage{floatrow}
\usepackage{multirow, booktabs}
\newfloatcommand{capbtabbox}{table}[][\FBwidth]

\usepackage{aaai21}  
\usepackage{times}  
\usepackage{helvet} 
\usepackage{courier}  
\usepackage[hyphens]{url}  
\usepackage{graphicx} 
\usepackage{natbib}  
\usepackage{caption} 
\title {mForms : Multimodal Form-Filling with Question Answering}

\author{
    Larry Heck, Simon Heck, Anirudh Sundar\\
}
\affiliations {
    Georgia Institute of Technology \\
         \{larryheck, sheck6, asundar34\}@gatech.edu}

\begin{document}

\maketitle

\begin{abstract}
This paper presents a new approach to form-filling by reformulating the task as {\em multimodal natural language Question Answering (QA)}. The reformulation is achieved by first translating the elements on the GUI form (text fields, buttons, icons, etc.) to natural language questions, where these questions capture the element's multimodal semantics. After a match is determined between the form element (Question) and the user utterance (Answer), the form element is filled through a pre-trained extractive QA system. By leveraging pre-trained QA models and not requiring form-specific training, this approach to form-filling is zero-shot. The paper also presents an approach to further refine the form-filling by using multi-task training to incorporate a potentially large number of successive tasks. Finally, the paper introduces a multimodal natural language form-filling dataset Multimodal Forms (\mbox{mForms}), as well as a multimodal extension of the popular ATIS dataset to support future research and experimentation. Results show the new approach not only maintains robust accuracy for sparse training conditions but achieves state-of-the-art F1 of 0.97 on ATIS with approximately 1/10th the training data. 
 \end{abstract}

\section{Introduction}
\label{sec:introduction}
\begin{figure}[t]
\begin{center}
\noindent
	\includegraphics[width=2.25in]{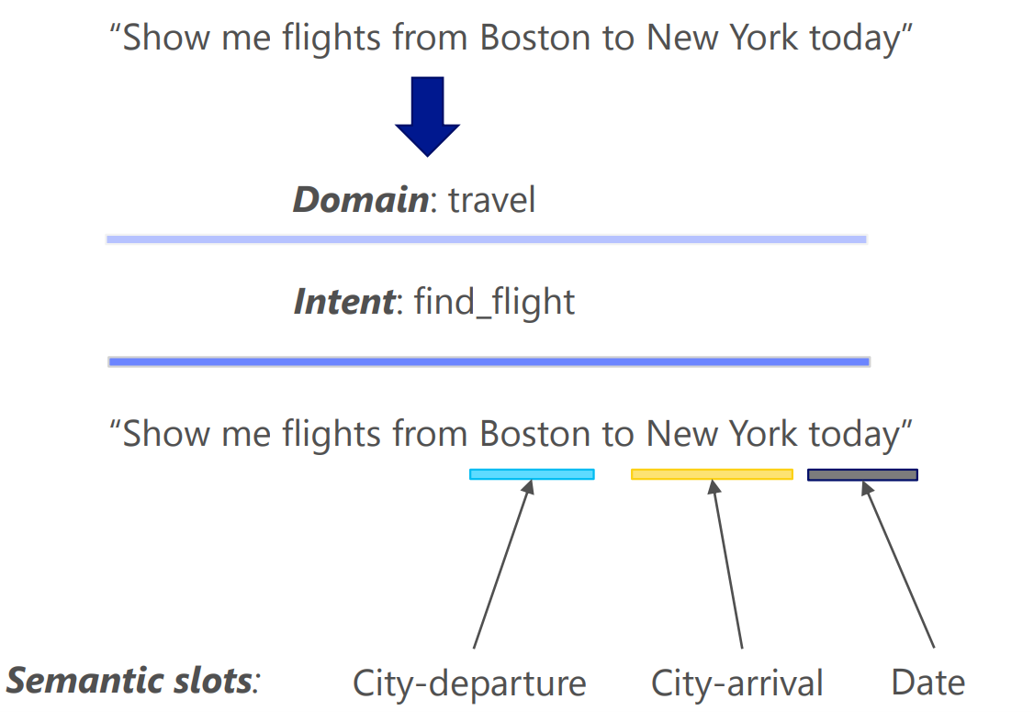}
	\caption{An example semantic representation with domain, intent, and semantic slot annotations.}\label{NLU}
\end{center}
\vspace*{-.25in}
\end{figure}

The last decade has seen the development and broad deployment of digital assistants (DAs) including Siri, Cortana, Alexa, Google Assistant, and Bixby. A primary component of DAs is Natural Language Understanding (NLU) - understanding the meaning of the user's utterance. Referring to Figure \ref{NLU}, the NLU task determines the domain of the user's request (e.g., travel), the user's intent (e.g., find\_flight) and information-bearing parameters commonly referred to as semantic slots (e.g., City-departure, City-arrival, and Date). The task of determining the semantic slots is called slot filling \cite{TurDeMori2011}. In this paper, we address a related but distinct task - \textit{form-filling}, where the DA processes the user requests to act on form elements (fill text fields, click buttons and icons, etc.) on Mobile Apps or web pages. Equipping DAs with the ability to simultaneously parse visual semantic information and contextual dialogue enhances their ability to understand and act on information across multiple modalities. This type of multimodal interaction through conversations is currently an open problem and an active area of research \cite{sundar2022multimodal}. 

Early methods for the related area of semantic slot filling used recurrent neural networks (RNNs) \cite{mesnil2014using}, then progressed to long short-term memory (LSTM) neural networks \cite{liu2016attentionbased}, and more recently transformer-based approaches \cite{chen2019bert}.

Dynamic deep learning approaches, while achieving high slot-filling accuracy, demand extensive domain-specific supervised training data. This poses challenges for applications with limited access to extensive data, such as AI skill development for DAs, limiting the broad expansion of AI skills to adequately cover the long tail of user goals and intents.

Prior work has focused on developing models and approaches that require less supervised training data. Zero and few-shot learning methods have been developed across NLP tasks \cite{dauphin2013zero,yann2014zero,upadhyay2018almost}. Methods can be broadly categorized into transfer learning \cite{jaech2016domain, el2014extending, hakkani-tr2016multi-domain}, sequential learning \cite{bapna2017sequential}, reinforcement learning \cite{liu2017end, kumar2017federated, shah2016} and synthetic training \cite{sileixu2020,giovanni2020}. 

In many cases, the user interacts with an App screen or web page and, therefore, uses multiple modalities such as voice, vision, and/or touch \cite{heck2013multi,hakkani2014eye,li2019rilod,selvaraju2019taking,zhang2020class, xu2021grounding, reichman-okvqa,zhang2019generative, sundar-heck-2023-ctbls,reichman-iccv23, sundar2024gtbls}. For these settings, zero- and few-shot learning can be achieved by leveraging the semantics contained in the screen. In \cite{bapna2017towards}, the authors incorporated visual slot names or descriptions in a domain-agnostic slot tagging model called a Concept Tagger. The Concept Tagger models the visual slot description (e.g. ``destination") as a Bag-of-Words (BOW) embedding vector and injects a Feed-Forward network inside the original deep LSTM network to process the user's utterance (e.g., ``Get a cab to 1945 Charleston"). Results showed the inclusion of slot descriptions significantly outperformed the previous state-of-the-art multi-task transfer learning approach \cite{hakkani-tr2016multi-domain}. 

The Concept Tagger \cite{bapna2017towards} is limited in several ways. First, the BOW semantic representation of the visual slot description is static and does not model the dynamics of the description language. Second, the method is limited to only visual slots with text descriptions and does not incorporate other semantic information from the visual elements (i.e., is the element a form field or a radio button with choices). Third, the Concept Tagger incorporates multi-task learning only through the visual slot description. 

This paper\footnote{Update of arXiv preprint \cite{mforms_arxiv}.} addresses all three limitations of the Concept Tagger. To address these limitations, the next section describes a new approach that formulates multimodal form-filling as Question Answering (QA). This approach also extends more recent work on text-based slot filling as QA \cite{LevySCZ17, du-etal-2021-qa,fuisz2022improved} by developing a much broader, multimodal computer vision-based approach. The extension to a multimodal approach is required in form-filling where the QA formulation must cover all of the 25 UI component categories, 197 text button concepts, and 99 icon classes. 

In the Experiments Section, we introduce a new corpus collected for multimodal form-filling called the Multimodal Forms (\mbox{mForms}) dataset as well as an extension of the ATIS \cite{tur2010left} dataset as a simulated form-filling task. We compare the new zero-shot multimodal form-filling QA approach to competing methods on this new corpora. Finally, we summarize our findings and suggest the next steps in the Conclusions and Future Work Section.

\section{Approach}
\label{approach}
\subsection{Multimodal form-filling}
The foundation of the approach presented in this paper is the utilization of deeper semantics {\em in the visual representation of the form} on the user's screen. While previous form-filling methods treated the form label as a classification tag with no semantic information, the approach of this paper extracts meaning from the visual slot representation. By formulating the form field description as a Question and the user's utterance as the Paragraph, we can directly utilize transformer-based extractive Question Answering (QA) models \cite{lan2019albert}. The Start/End Span of the extracted Answer is used to fill the appropriate content in the web form. We call our approach {\em Multimodal form-filling as Question Answering (QA)}, which we will henceforth refer to by {\em \mbox{mForms} as QA}.


\begin{figure}[t]
\begin{center}
\noindent
	\includegraphics[width=3.25in]{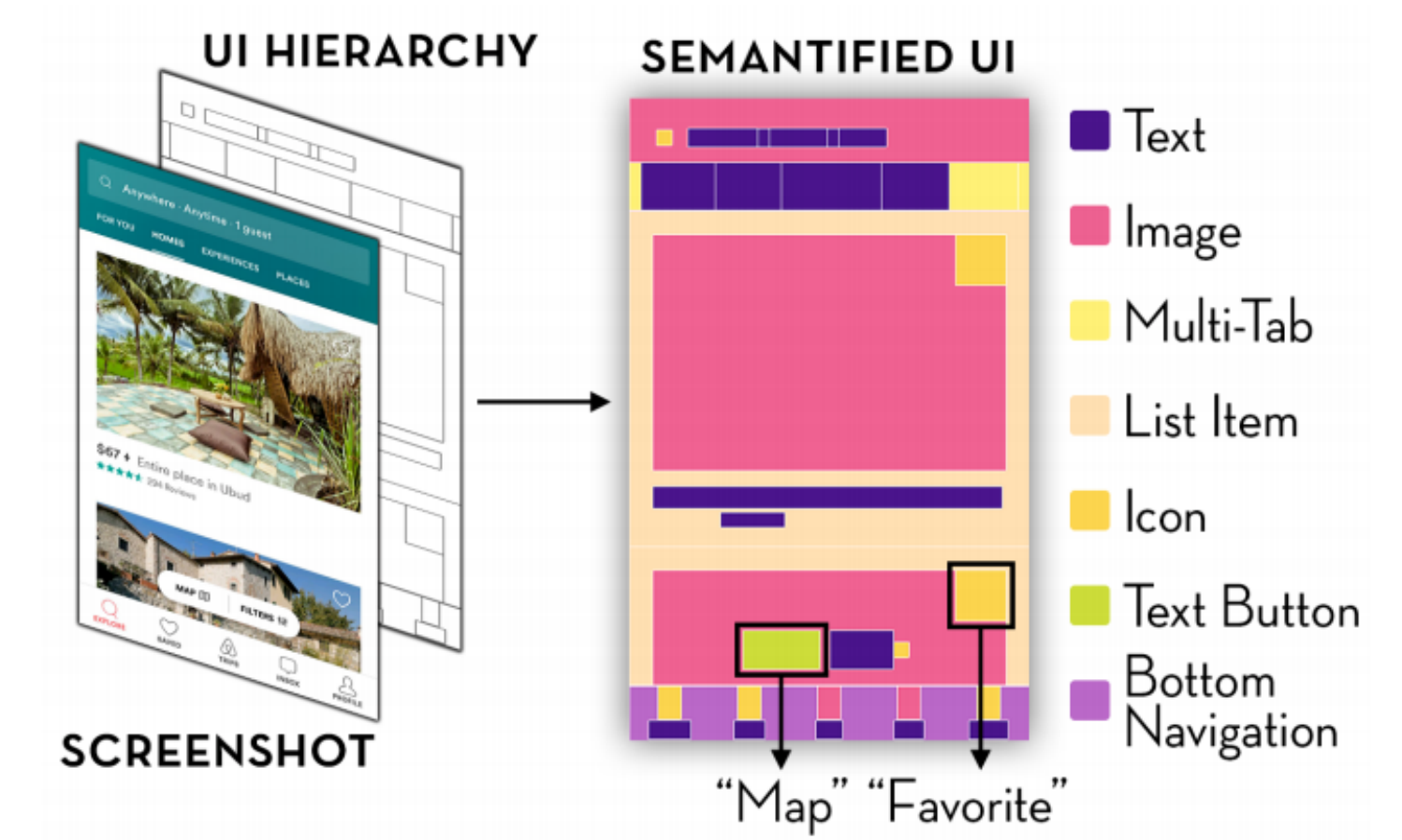}
	\caption{Semantically annotated mobile Graphical User Interface (GUI) using computer vision to identify 25 UI component categories, 197 text button concepts, and 99 icon classes (Liu et al. 2018)}\label{SemanticApp}
\end{center}
\vspace*{-.35in}
\end{figure}

In addition to the lexical semantics contained in the text field description, the type of the visual graphical user interface (GUI) element on the App or web page provides additional semantic information. The set of GUI design elements of a mobile App that are available to translate into questions are shown in Figure \ref{SemanticApp}. In our approach, the GUI design elements are automatically classified via a convolutional deep neural network computer vision system trained on the RICO dataset as shown in Figure \ref{GUI_CV} of the Appendix \cite{Deka:2017:Rico,Liu:2018}. The computer vision classifier identifies 25 UI component categories (e.g., Ads, Checkboxes, On/Off switches, Radio Buttons), 197 text button concepts (e.g., login, back, add/delete/save/continue), and 99 icon classes (e.g., forward/backward, dots, plus symbol). Our implementation as described in \cite{Liu:2018} has a 94\% classification accuracy. 

\begin{figure*}[t]
    \centering
    \includegraphics[width=\textwidth]{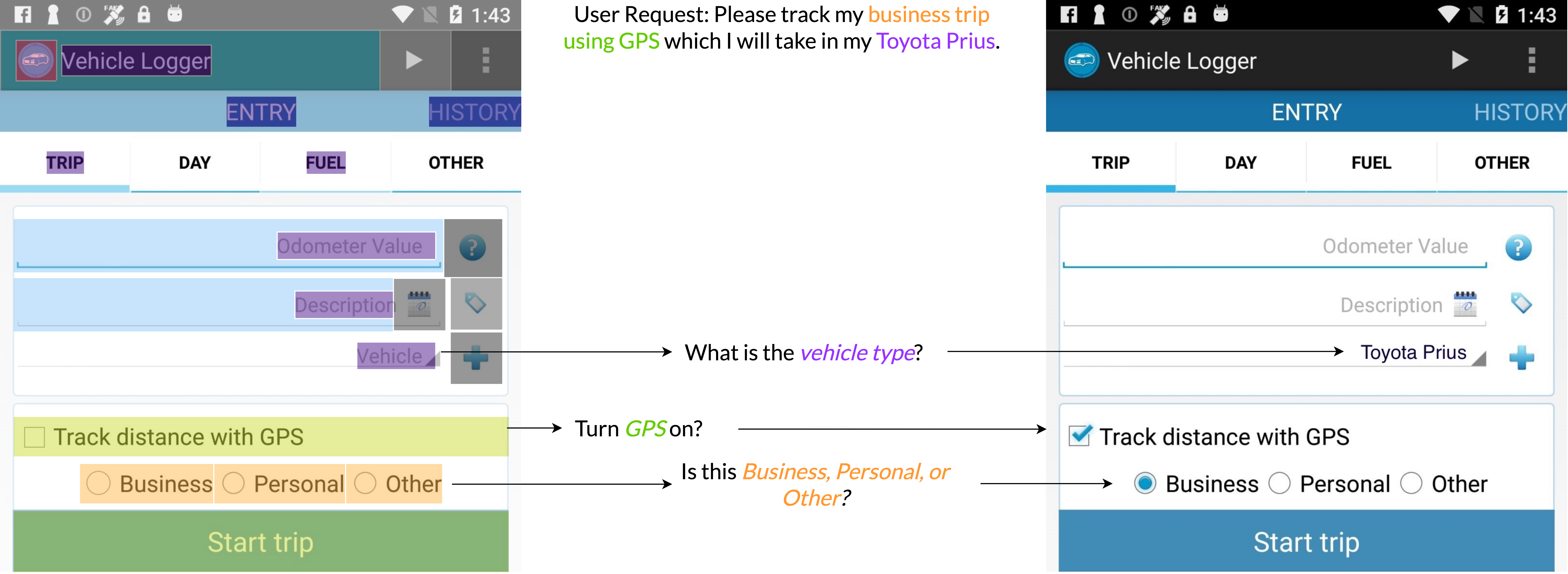}
    \caption{The mForms pipeline. The rule template uses the semantified UI to trigger a question template. The visual information of the GUI element drives the generation of the actual question. Then, using the user's request as evidence, the questions are answered to fill the form with the appropriate information. }
    \label{fig:GUI_to_QA}
\end{figure*}

In \mbox{mForms}, rules are used to translate each GUI design element into an appropriate question. Each type of GUI design element has a unique rule type that triggers depending on its visual presence on the GUI. Figure \ref{fig:GUI_to_QA} shows an example of these GUI elements, their associated rule templates, and example questions and user utterances. If multiple GUI design elements are visible, then multiple translation rules fire, generating simultaneous questions to be paired with the user's utterance.


For example, GUI elements that are classified as simple text fields trigger a rule that generates a question template ``What is the \textit{Text\_Field}?". Figure \ref{fig:GUI_to_QA} shows simple text fields in the Michaelsoft Vehicle Logger App. The first text field is ``Vehicle". In this case, the rule recognizes command and generates a question ``What is the vehicle type?". Given a user utterance ``Please track my business trip using GPS which I will take in my Toyota Prius.", the Question-Answering system extracts the {\em answer} to this question as ``Toyota Prius". 

\subsection{Single- and Multi-Task Training}
\begin{figure}[h]
\begin{center}
\noindent
\hspace{-.2in}
	\includegraphics[width=3.2in]{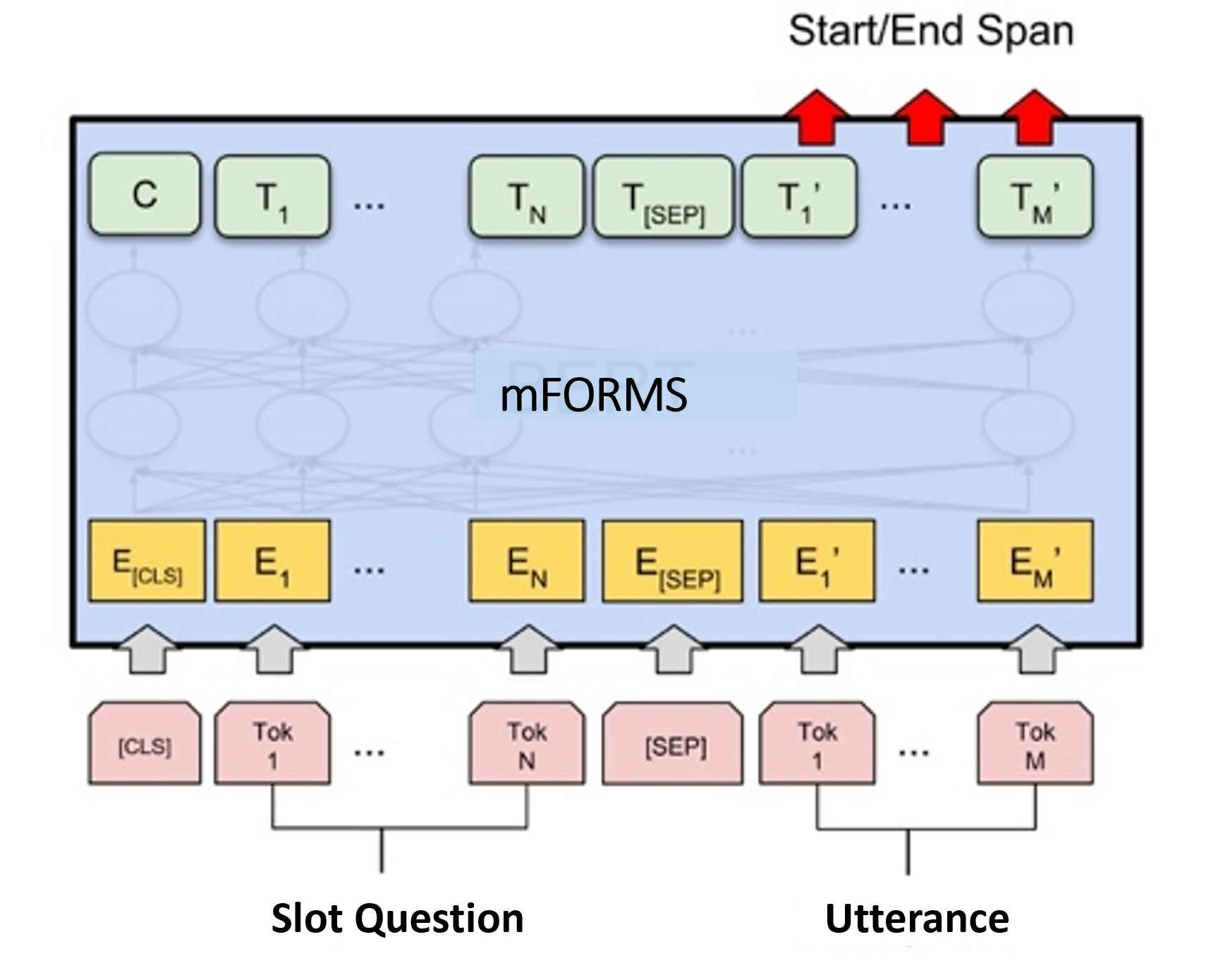}
	\caption{mForms as QA Approach}\label{mFormsApproach}
\end{center}
\end{figure}

Our \mbox{mForms as QA} method is shown in Figure \ref{mFormsApproach}. It can be formulated as both single- and multi-task training.  
The Single-task (ST) model is initialized as a general-purpose QA trained with SQuAD2 \cite{rajpurkar2018know}. Used as-is, this model is {\it zero-shot} for form-filling. The model can be fine-tuned with supervised (annotated) form-filling data from the visual App or web page GUI.  

In contrast to Single-task training, Multi-task (MT) training incorporates form-filling training sets across multiple tasks with each training further refining the model. Similar tasks represented by common domains can be grouped for successive fine-tuning stages. For example, flight reservation form-filling Apps could be successively refined using the first N$-$1 Apps with the Nth App used as the final fine-tuning stage. The potential advantage of the MT approach is the required amount of annotated supervised training data becomes less with each new task refinement stage. 

\section{Experiments}
\subsection{Setup}
Our base QA system is based on the Pytorch implementation of ALBERT \cite{lan2019albert} \footnote{https://github.com/huggingface/transformers}. We use the pre-trained LM weights in the encoder module trained with all the official hyper-parameters\footnote{ALBERT (xxlarge)}.

\subsection{Multimodal Forms Dataset} 
Amazon Mechanical Turk (AMT) was used to collect Multimodal Forms (\mbox{mForms}) - a dataset to support multimodal form-filling research\footnote{https://huggingface.co/avalab}. The AMT crowd workers were asked to formulate requests to mobile App screens from three Apps in the RICO dataset: Vehicle Logger from Michaelsoft, United Airlines flight search, and Trip Advisor. The UIs of each App with GUI elements semantically annotated by the computer vision system described earlier are available online.\footnote{http://interactionmining.org/rico} More details on the \mbox{mForms} dataset are in given in the Appendix.

\begin{table*}[t]
\centering
\begin{small}
\begin{tabular}{l|l|c}
\toprule
     Visual App    &       Sample Utterance                                                 & \# Utterances \\
\midrule
Vehicle Logger    & ``Please activate GPS tracking and log my car trip" & 850 \\ 
United    & ``Book a Flight from California to arizona on august 15th 2020" & 850 \\ 
ATIS form-filling    & ``I live in Denver and I'd like to make a trip to Pittsburgh" & 4478 \\ 
Trip Advisor    & ``Please book a 5 star hotel in Atlanta Georgia"  & 803 \\ 
\bottomrule
\end{tabular}
\end{small}
\caption{Sample utterances from each domain} 
\label{sample_utts_datasets}
\end{table*}

\subsection{Simulated ATIS Form-Filling Dataset}
\begin{figure}[h]
\begin{center}
\noindent
	\includegraphics[width=3.2in]{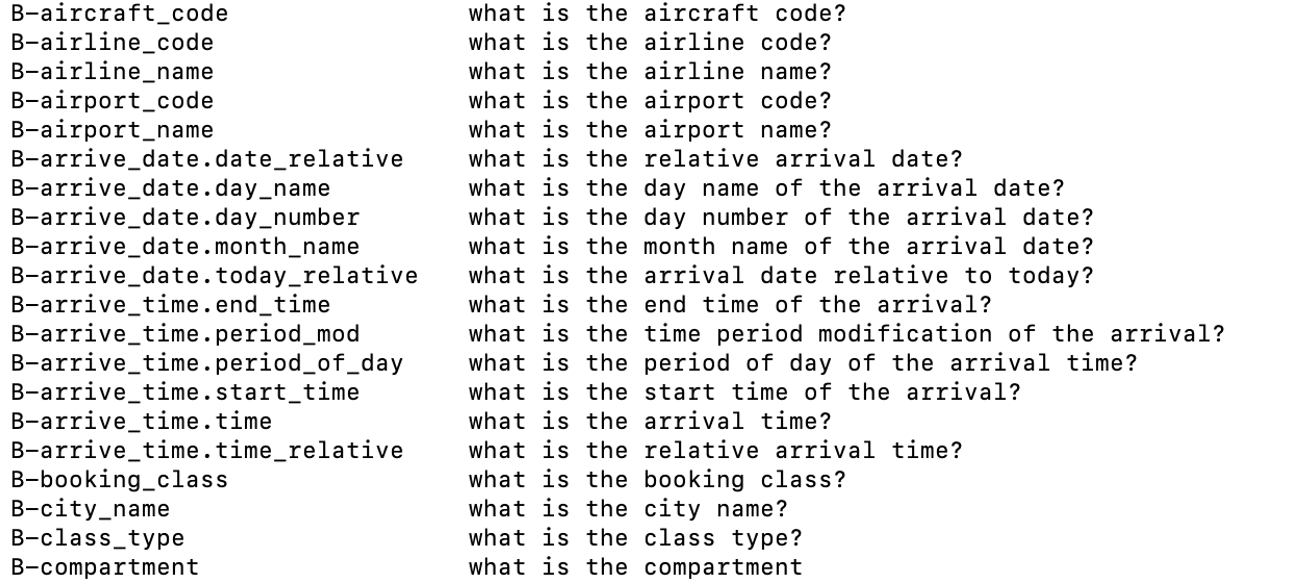}
	\caption{Simulated ATIS form-filling fields translated to Questions}\label{ATISdesc}
\end{center}
\vspace*{-.1in}
\end{figure}

The ATIS dataset is a widely used NLU benchmark for users interacting through natural language with a flight booking system \cite{tur2010left}. To use ATIS for mForms as QA, we extended the dataset in several ways. First, as shown in Figure \ref{ATISdesc}, each slot is treated as a simulated visual Text Field where the information of the slot tag is displayed in an App with a simple form. As is the case with Text Fields, each slot was reformulated as a natural language Question. For example, the ATIS slot tag ``B-aircraft\_code" is translated into the question ``What is the aircraft code?". This modified dataset will be called ``ATIS form-filling".

Table \ref{sample_utts_datasets} summarizes the three Visual App datasets as well as ATIS form-filling with example utterances from each dataset. Table \ref{slot schema} in the Appendix shows the types of slots annotated in each dataset. ATIS form-filling has the largest number of slot types at 83.

\begin{table*}[h]
\centering
\begin{small}
\begin{tabular}{l|c c | c c | c c | c c |c c}
\toprule
\# train samples & \multicolumn{2}{c|}{0}   & \multicolumn{2}{c|}{5}    & \multicolumn{2}{c|}{50}   & \multicolumn{2}{c|}{100}  & \multicolumn{2}{c}{500} \\ \hline
Domain          & JB    & mForms       &  JB   & mForms       & JB    & mForms       &  JB       & mForms   &  JB   & mForms\\
\midrule
Vehicle Logger  &0.00   & 0.48              &0.00   & 0.46                 & 0.48  &0.73                  &0.45       &0.80           &0.78   &0.87      \\ 
ATIS form-filling    &0.00   & 0.60              &0.00   &0.74               &0.66   & 0.88              & 0.77      &0.93           &0.91   &0.97       \\
United          &0.00   &0.40               &0.00   &0.44               &0.37   & 0.58              &0.44       &0.72           &0.51   &0.74       \\
Trip Advisor   &0.00    &0.52               &0.00   &0.47               &0.18    &0.63               &0.53      &0.66           &0.59   &0.66       \\
\bottomrule
\end{tabular}
\end{small}
\caption{Weighted token F1 (harmonic mean of precision and recall) scores. The table shows the baseline (JB) as detailed in \cite{chen2019bert} versus our new mForms as QA approach.} 
\label{overall_results}
\end{table*}

Table \ref{overall_results} summarizes F1 scores (harmonic mean of precision and recall) on the 3 new mForms datasets and the ATIS form-filling dataset. For comparison, the F1 score is given for the joint slot and intent model (JB) given in \cite{chen2019bert}. The new mForms as QA approach presented in this paper consistently outperforms the JB slot filler. While the JB slot filler requires at least 100 training samples on the Vehicle Logger App, the mForms as QA approach maintains the F1 score even for only 0 and 5 training samples. These results suggest the semantic information contained in the mForms is particularly important for sparse training conditions.

With more training data, is interesting to note that the accuracy of the new mForms as QA approach also achieves one of the best published F1 measures at 0.97 on the ATIS dataset. For comparison, the mForms as QA system was only trained on 500 samples for this case as compared to the full training set of ATIS at over 4400 samples. This suggests that the injection of simulated mForm and the subsequent generation of questions for the QA system is effective at reducing the amount of training data required to yield high accuracy.  This characteristic of mForms as QA makes the approach especially attractive for commercial digital assistants given the industry's reliance on third-party developers who are often not highly skilled in NLU.

To examine the effect of visual semantics in a more controlled experiment, questions generated in the new mForms as QA approach were replaced with tag symbols, where the tag symbol had no semantic information (e.g., ``XYZ"). Otherwise, ``No Visuals" is the same model as mForms as QA. Results for the Vehicle Logger App are shown in Table \ref{visual_semantics_results} comparing the No Visuals approach to two conditions from the mForms as QA approach (1) Text Only - all visuals are treated as simple Text Fields and other GUI elements are ignored (2) All GUI elements are used. The training samples were randomly chosen across all 10 slot types from the complete set of 500 utterances. Larger differences are observed in sparse training conditions where the No Visuals approach largely falters.  
\begin{table}[h]
\centering
\begin{small}
\begin{tabular}{l|c|c|c|c}
\toprule
\# train samples    & 0     & 50    & 100   & 500\\ 
\midrule
No Visuals          & 0.01  & 0.29  & 0.32  & 0.71 \\ 
Text Visuals        & 0.36  & 0.69  & 0.71     & 0.88 \\ 
GUI (all) Visuals   & 0.48  & 0.73  & 0.80  & 0.87 \\ 
\bottomrule
\end{tabular}
\end{small}
\caption{F1 results showing effects of visual semantics on the Vehicle Logger App. The row labeled Text Visuals shows the results of our mForms as QA method with every visual element treated as a simple text field. GUI (all) Visuals leverage the full semantic information contained in the visual GUI elements for mForms as QA.} 
\label{visual_semantics_results}
\end{table}

Given the mForms as QA approach incorporates multi-task (MT) training, an interesting question to answer is whether the MT training transfers knowledge \emph{across domains}. Table \ref{horizontal_MT_results} shows results for the cross-domain case: fine-tuning on the ATIS form-filling dataset followed by another iteration of fine-tuning on data from the Vehicle Logger App. The effect of MT is more pronounced in the sparse training cases with an improvement from 0.48 F1 to 0.52 F1 at zero-shot training and an improvement of 0.46 F1 to 0.60 F1 with 5 training samples. These results suggests cross-domain concept learning is occurring for these training conditions. 
\begin{table}[h]
\centering
\begin{small}
\begin{tabular}{l|c|c|c|c}
\toprule
\# train samples        & 0             & 5         & 100       & 500     \\
\midrule
Vehicle Logger                      & 0.48          & 0.46         &0.80       &0.87     \\ 
+ATIS form-filling      & 0.52          &0.60       &0.80       &0.89 \\
\bottomrule
\end{tabular}
\end{small}
\caption{Results for multi-task training (MT) across multiple domains. The first row shows F1 scores for the Vehicle Logger dataset for various amounts of training data. The second row shows the effect of fine-training the SQuAD2 model with ATIS form-filling before training with the Vehicle Logger data.} 
\label{horizontal_MT_results}
\end{table}

Finally, Table \ref{visualelements_results} shows zero-shot F1 scores on the mForms datasets for our new approach when varying the number of visual GUI elements that are displayed to the user. For example, when 2 visual elements are displayed, the model must not only parse the slots from the utterance for one of the visual elements but also correctly reject the filling of slots into the other element. For the mForms datasets, the models degrade gracefully. This robustness is likely the result of the initial fine-tuning on the SQuAD2 dataset which is trained to reject false questions - questions that do not have a correct answer to extract from the given Paragraph. 
\begin{table}[h]
\hspace*{-.1in}
\begin{small}
\begin{tabular}{l|c|c|c|c|c}
\toprule
\# elements  & 1     & 2     & 3     & 4     & 5 \\
\midrule
Vehicle Logger          &0.52   &0.51   &0.49   &0.49   &0.46    \\
ATIS form-filling            & 0.60      &  0.58     &  0.56 & 0.53  &  0.52 \\
\bottomrule
\end{tabular}
\end{small}
\caption{Zero-Shot Slot F1 scores on the Vehicle Logger and ATIS form-filling datasets for varying numbers of visual elements shown to the user simultaneously}  
\label{visualelements_results}
\end{table}

\section{Conclusions and Future Work}
This paper presented a new approach to filling GUI forms by reformulating the problem as a multimodal natural language
Question Answering (QA) task.
The reformulation is achieved by first translating the elements on the GUI form (text
fields, buttons, icons, etc.) to natural language questions, where these questions capture the element’s multimodal
semantics. These questions are paired with the user's utterance and answers are extracted from the utterance using a Transformer-based Question-Answering system. An approach to further refine the model is presented that facilitates transfer learning across a large number of tasks. 
The paper also presented \mbox{mForms} - a multimodal form-filling dataset to support future research.

Results show the new approach not only maintains robust accuracy for sparse training conditions but achieves state-of-the-art F1 of 0.97 on ATIS with approximately 1/10th of the training data. 

Future work will extend \mbox{mForms as QA} to a broader set of visual GUI screens across both mobile Apps and web pages. In addition, we plan to explore improved rejection methods for screens with high-density competing visual GUI elements. Lastly, while \mbox{mForms as QA} uses a BERT-based architecture for comparison with prior work, future work will explore ways to leverage generative models such as GPT3.5-4/T5/BART.

\vspace*{.2in}

\bibliography{nlu_bibliography}


\newpage
\vspace*{8in}
\newpage
\section{Appendix}
\label{sec:appendix}

This section outlines the applications used to collect the \mbox{mForms} dataset. The slot schemas for each form are shown in Table \ref{slot schema}.\\

\begin{figure}[h]
\begin{center}
\noindent
\hspace{-.1in}
	\includegraphics[width=2in]{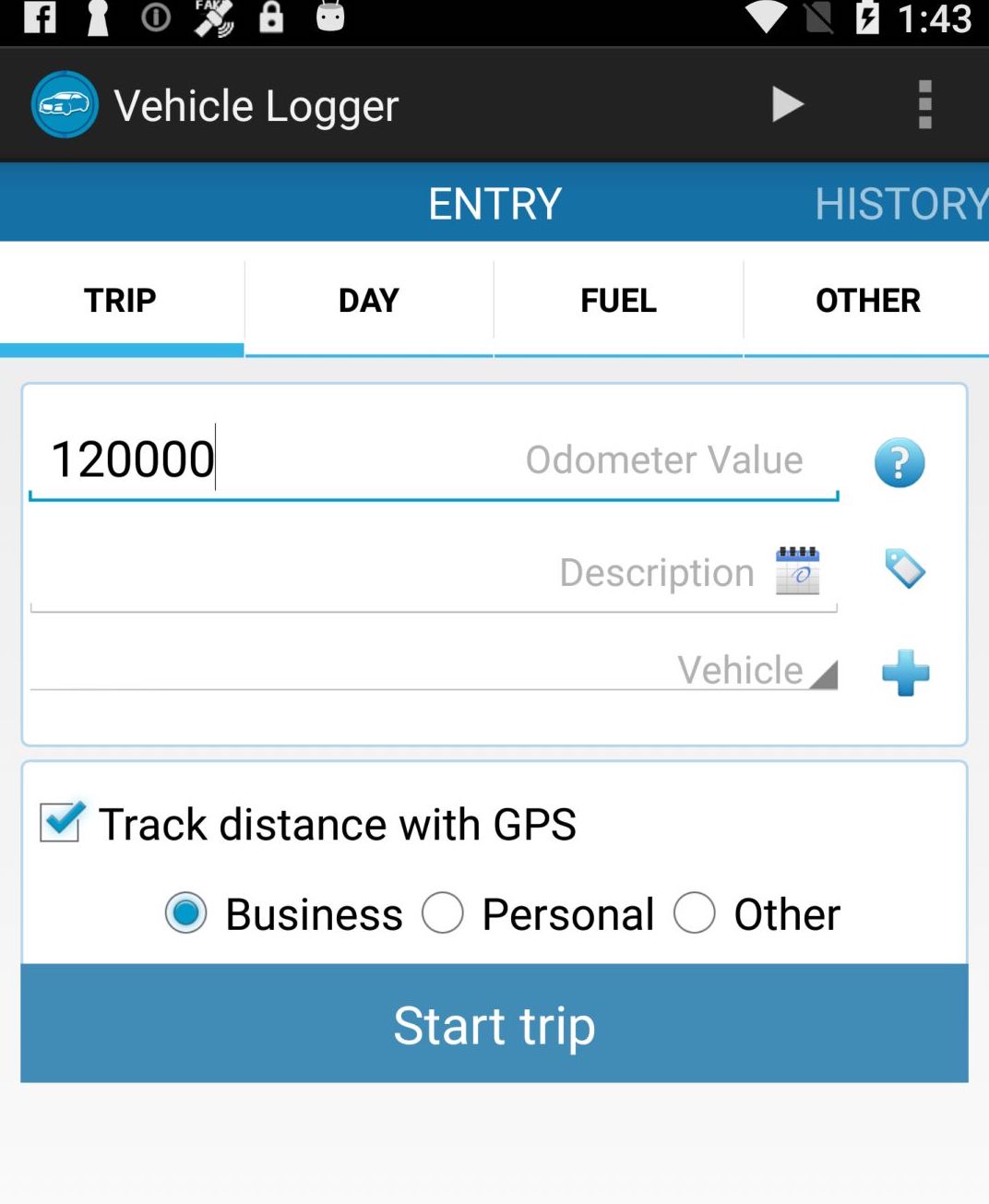}
	\caption{Vehicle Logger Application}\label{fig:tripLoggerApp}
\end{center}
\end{figure}

\textbf{Vehicle Logger:} The Vehicle Logger App shown in Figure \ref{fig:tripLoggerApp} is a popular tool to create, share, and report vehicle log books for mileage, fuel expenses, and tax purposes. As previously described, the visual GUI elements of the Vehicle Logger App include Text fields (e.g., Odometer Value), Radio Buttons (e.g., Business, Personal, Other), and Text Buttons (e.g., Track distance with GPS). Referring to Table \ref{sample_utts_datasets}, 850 utterances were collected with annotations according to 10 slot types. \\

\begin{figure}
\vspace*{-0.5in}
\begin{center}
\noindent
\hspace{-.1in}
	\includegraphics[width=2.3in]{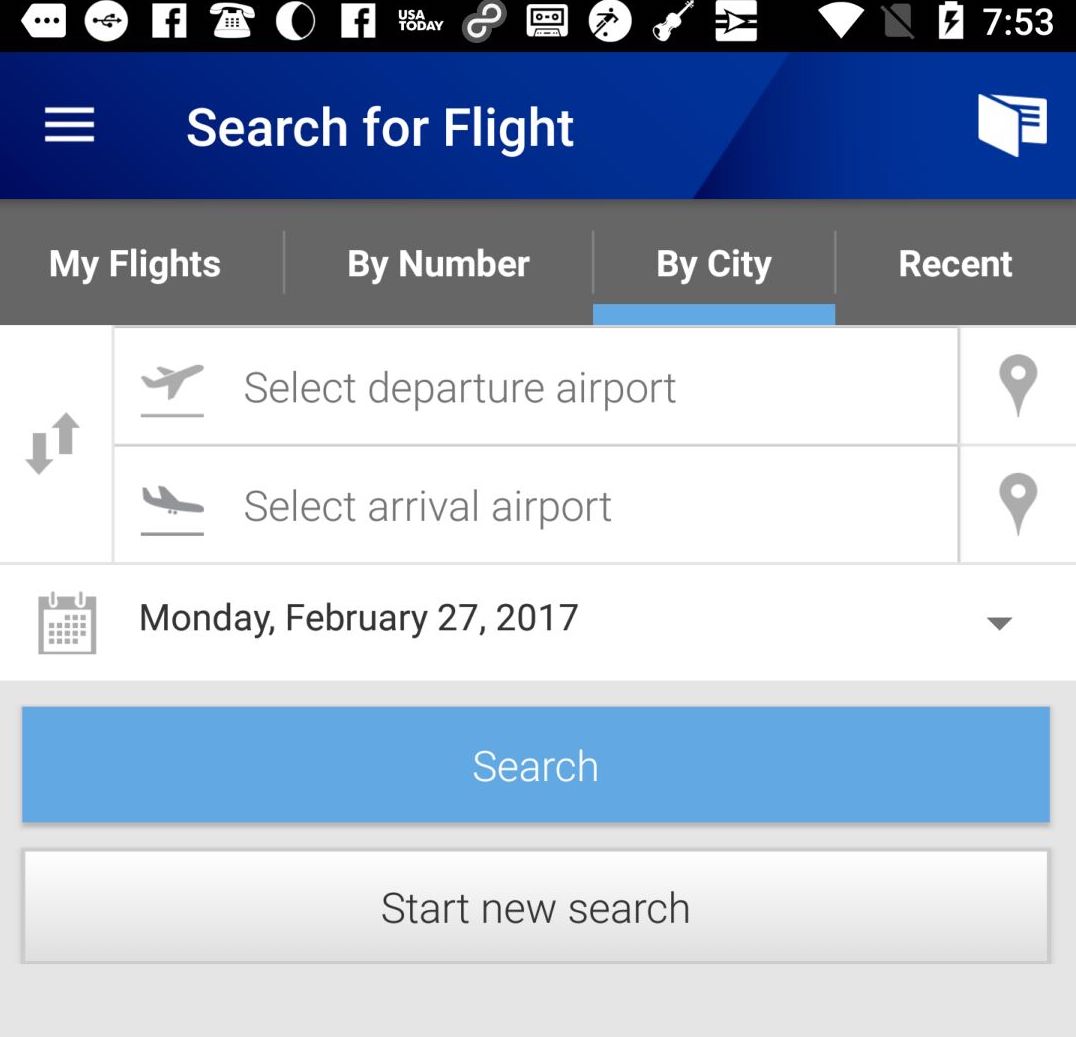}
	\caption{United flight search Application}\label{fig:unitedApp}
\end{center}
\end{figure}

\textbf{United Airlines: } The United Airlines flight search App shown in Figure \ref{fig:unitedApp} is used to find flights according to travel plans and preferences. The GUI elements include simple text fields, tab buttons, and search buttons as well as more visually-oriented icons such as the user's current location (icons on the right-most column) and an icon to swap departure and arrival airports. 850 utterances were collected with annotations according to 6 slot types. \\

\begin{figure}
\vspace*{-0.2in}
\begin{center}
\noindent
\hspace{-.1in}
	\includegraphics[width=2.3in]{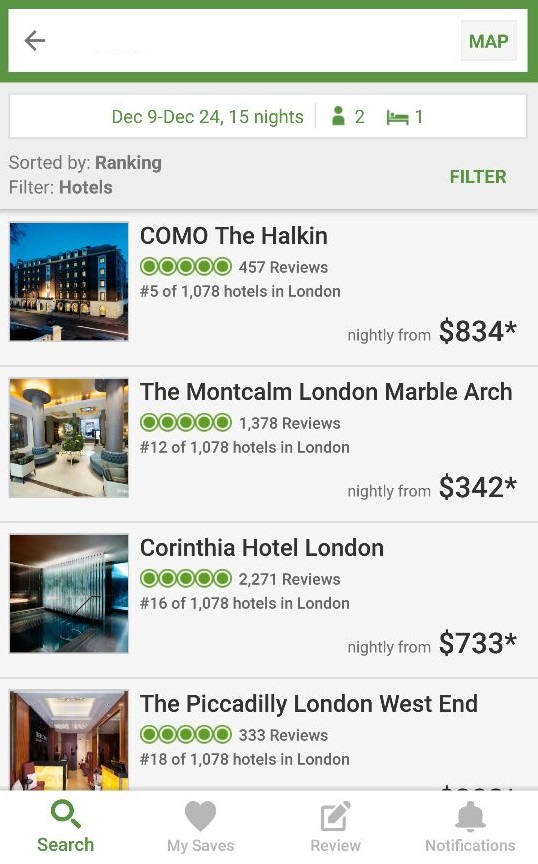}
	\caption{Trip Advisor Application}\label{fig:tripAdvisorApp}
\end{center}
\end{figure}

\textbf{Trip Advisor: } Finally, Figure \ref{fig:tripAdvisorApp} shows the Trip Advisor App. This App serves many purposes including booking a table at restaurants as well as comparing prices when booking flights and hotels. The portion of the App used for this study focused on hotel room booking. Much of the App screen shown in the Figure contains visually oriented icons such as the symbol for people (in this case, showing 2 people) and a bed (1 bed in the room). The Trip Advisor dataset has 803 utterances with annotations according to 6 slots.

\begin{table*}
\centering
\noindent
\hspace{-.1in}
\begin{small}
\begin{tabular}{l|l}
\toprule
     Visual App    &       Slot descriptions             \\
\midrule
\multirow{2}{*}{Vehicle Logger}    & fuel cost, fuel added, trip description, gps tracking, start logging, date, odometer value, \\
& trip type, entry, vehicle \\ \hline
United    &  arrival airport, departure airport, travel dates, search, switch/swap airports\\ \hline
\multirow{2}{*}{ATIS form-filling}    & aircraft code, airline code, airline name, airport code, airport name, arrival date (relative), \\
                & arrival date (day name), arrival date (day number), arrival date (month name), etc\\ \hline
Trip Advisor    &  number of beds, date range, filter by price, filter by rating, number of nights, number of people\\ 
\bottomrule
\end{tabular}
\end{small}
\caption{Slot schema / descriptions used for the mForms tagger for each domain} 
\label{slot schema}
\end{table*}

\begin{figure*}
\vspace*{-2in}
\begin{center}
\noindent
	\includegraphics[width=7in]{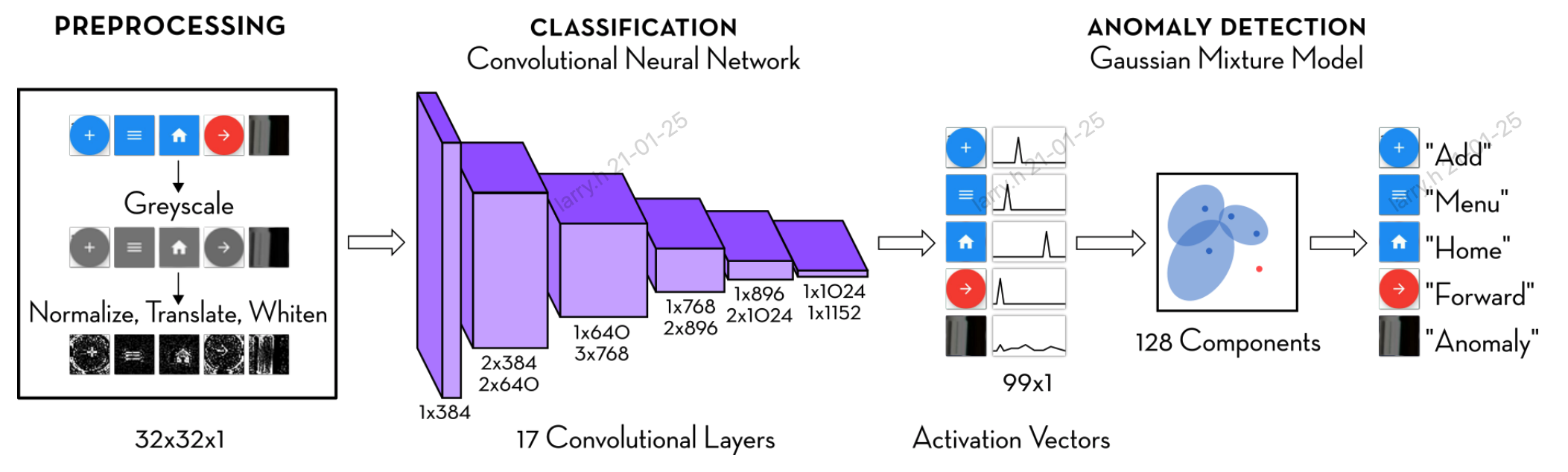}
	\caption{Computer Vision classification of GUI visual elements (Liu et al. 2018)}\label{GUI_CV}
\end{center}
\vspace*{-.2in}
\end{figure*}


\end{document}